\documentclass[11pt,a4paper]{article}
\pdfoutput=1

\usepackage[hyperref]{emnlp2018}
\usepackage{times}
\usepackage{latexsym}
\usepackage{url}
\usepackage{subfig}
\usepackage{color}
\usepackage{graphicx}
\usepackage{array,multirow,multicol}
\usepackage{epstopdf}
\usepackage{amssymb}
\usepackage{amsmath}
\usepackage{enumitem}
\usepackage{tabularx}
\usepackage{setspace}
\usepackage{soul}

\newcommand{\softmax}{\operatornamewithlimits{softmax}}

\newcommand{\milnet}{\textsc{MilNet}}
\newcolumntype{Y}{>{\centering\arraybackslash}X}

\aclfinalcopy % Uncomment this line for the final submission

\title{Summarizing Opinions: Aspect Extraction Meets
  Sentiment Prediction and They Are Both Weakly Supervised}

\author{Stefanos Angelidis \textnormal{and} Mirella Lapata \\
   Institute for Language, Cognition and Computation \\
   School of Informatics, University of Edinburgh \\
   10 Crichton Street, Edinburgh EH8 9AB \\
   {\tt s.angelidis@ed.ac.uk, mlap@inf.ed.ac.uk} \\\\ 
}

\date{}

\begin{document}
\maketitle
\begin{abstract}
  We present a neural framework for opinion summarization from online
  product reviews which is knowledge-lean and only requires light
  supervision (e.g., in the form of product domain labels and
  user-provided ratings). Our method combines two weakly supervised
  components to identify salient opinions and form extractive
  summaries from multiple reviews: an aspect extractor trained under a
  multi-task objective, and a sentiment predictor based on multiple
  instance learning. We introduce an opinion summarization dataset
  that includes a training set of product reviews from six diverse
  domains and human-annotated development and test sets with gold
  standard aspect annotations, salience labels, and opinion summaries.
  Automatic evaluation shows significant improvements over baselines,
  and a large-scale study indicates that our opinion summaries are
  preferred by human judges according to multiple
  criteria.\footnote{Our code and dataset are publicly available at
  \url{https://github.com/stangelid/oposum}.}
\end{abstract}

\section{Introduction}
\label{sec:intro}

Opinion summarization, i.e.,~the aggregation of user opinions as
expressed in online reviews, blogs, internet forums, or social media,
has drawn much attention in recent years due to its potential for
various information access applications. For example, consumers have
to wade through many product reviews in order to make an informed
decision. The ability to summarize these reviews succinctly would
allow customers to efficiently absorb large amounts of opinionated
text and manufacturers to keep track of what customers think about
their products \cite{liu2012sentiment}.

\begin{figure*}[t]
  \centering
  \includegraphics[width=\textwidth]{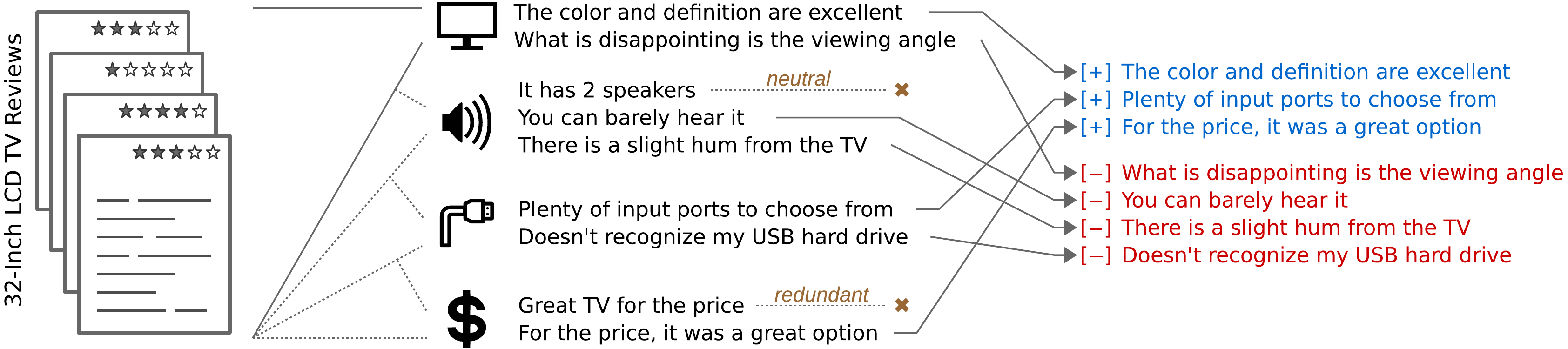}
  \caption{Aspect-based opinion summarization. Opinions on image
    quality, sound quality, connectivity, and price of an LCD
    television are extracted from a set of reviews. Their polarities
    are then used to sort them into positive and negative, while
    neutral or redundant comments are discarded.}
  \label{fig:task}
  \vspace{-1.5mm}
\end{figure*}

The majority of work on opinion summarization is
\emph{entity-centric}, aiming to create summaries from text
collections that are relevant to a particular entity of interest,
e.g.,~product, person, company, and so on. A popular decomposition of
the problem involves three subtasks \cite{Hu:Liu2004,Hu:Liu:2006}:
(1)~\emph{aspect extraction} which aims to find specific features
pertaining to the entity of interest (e.g.,~battery life, sound
quality, ease of use) and identify expressions that discuss them;
(2)~\emph{sentiment prediction} which determines the sentiment
orientation (positive or negative) on the aspects found in the first
step, and (3)~\emph{summary generation} which presents the identified
opinions to the user (see Figure~\ref{fig:task} for an illustration of
the task).

A number of techniques have been proposed for aspect discovery using
part of speech tagging \cite{Hu:Liu2004}, syntactic parsing
\cite{lu2009rated}, clustering \cite{Mei:ea:2007,titov2008modeling},
data mining \cite{Ku:ea:2006}, and information extraction
\cite{popescu-etzioni:2005:HLTEMNLP}. Various lexicon and rule-based
methods \cite{Hu:Liu2004,Ku:ea:2006,google:sum:2008} have been adopted
for sentiment prediction together with a few learning approaches
\cite{lu2009rated,pappas2017explicit,angelidis18milnet}. As for the
summaries, a common format involves a list of aspects and the number
of positive and negative opinions for each \cite{Hu:Liu2004}.  While
this format gives an overall idea of people's opinion, reading the
actual text might be necessary to gain a better understanding of
specific details. Textual summaries are created following mostly
extractive methods (but see \citealt{ganesan2010opinosis} for an
abstractive approach), and various formats ranging from lists of words
\cite{popescu-etzioni:2005:HLTEMNLP}, to phrases \cite{lu2009rated},
and sentences
\cite{Mei:ea:2007,google:sum:2008,lerman2009sentiment,wang2016neural}.

In this paper, we present a neural framework for opinion extraction
from product reviews. We follow the standard architecture for
aspect-based summarization, while taking advantage of the success of
neural network models in learning continuous features without recourse
to preprocessing tools or linguistic annotations. Central to our
system is the ability to accurately identify aspect-specific opinions
by using different sources of information freely available with
product reviews (product domain labels, user ratings) and
minimal domain knowledge (essentially a few aspect-denoting keywords).
We incorporate these ideas into a recently proposed aspect discovery
model \cite{he2017unsupervised} which we combine with a weakly
supervised sentiment predictor \cite{angelidis18milnet} to identify
highly salient opinions.  Our system outputs extractive summaries
using a greedy algorithm to minimize redundancy. Our approach takes
advantage of weak supervision signals only, requires minimal human
intervention and no gold-standard salience labels or summaries for
training.

Our contributions in this work are three-fold: a novel neural
framework for the identification and extraction of salient customer
opinions that combines aspect and sentiment information and does not
require unrealistic amounts of supervision; the introduction of an
opinion summarization dataset which consists of Amazon reviews from
six product domains, and includes development and test sets with gold
standard aspect annotations, salience labels, and multi-document
extractive summaries; a large-scale user study on the quality of the
final summaries paired with automatic evaluations for each stage in
the summarization pipeline (aspects, extraction accuracy, final
summaries).  Experimental results demonstrate that our approach
outperforms strong baselines in terms of opinion extraction accuracy
and similarity to gold standard summaries. Human evaluation further
shows that our summaries are preferred over comparison systems across
multiple criteria.

\section{Related Work}
\label{sec:related}

It is outside the scope of this paper to provide a detailed treatment
of the vast literature on opinion summarization and related tasks. For
a comprehensive overview of non-neural methods we refer the interested
reader to \citet{kim2011comprehensive} and \citet{liu2012survey}.  We
are not aware of previous studies which propose a neural-based
system for end-to-end opinion summarization without direct
supervision, although as we discuss below, recent efforts tackle
various subtasks independently.

\paragraph{Aspect Extraction}
Several neural network models have been developed for the
identification of aspects (e.g., words or phrases) expressed in
opinions. This is commonly viewed as a supervised sequence labeling
task; \citet{liu2015fine} employ recurrent neural networks, whereas
\citet{yin2016unsupervised} use dependency-based embeddings as
features in a Conditional Random Field (CRF).
\citet{wang2016recursive} combine a recursive neural network with CRFs
to jointly model aspect and sentiment terms.  \citet{he2017unsupervised}
propose an aspect-based autoencoder to discover fine-grained aspects
without supervision, in a process similar to topic modeling. Their
model outperforms LDA-style approaches and forms the basis of our
aspect extractor.

\paragraph{Sentiment Prediction}
Fully-supervised approaches based on neural networks have achieved
impressive results on fine-grained sentiment classification
\cite{kim2014convolutional,socher2013recursive}. More recently,
\textit{Multiple Instance Learning} (MIL) models have been proposed
that use freely available review ratings to train segment-level
predictors. \citet{kotzias2015group} and \citet{pappas2017explicit}
train sentence-level predictors under a MIL objective, while our
previous work \citep{angelidis18milnet} introduced \textsc{MilNet}, a
hierarchical model that is trained end-to-end on document labels and
produces polarity-based opinion summaries of single reviews. Here, we use
\textsc{MilNet} to predict the sentiment polarity of individual
opinions.

\paragraph{Multi-document Summarization}  
A few extractive neural models have been recently applied to generic
multi-document summarization. \citet{cao2015ranking} train a recursive
neural network using a ranking objective to identify salient
sentences, while follow-up work \cite{cao2017improving} employs a
multi-task objective to improve sentence extraction, an idea we
adapted to our task. \citet{yasunaga2017graph} propose a graph
convolution network to represent sentence relations and estimate
sentence salience. Our summarization method is tailored to the opinion
extraction task, it identifies aspect-specific and salient units,
while minimizing the redundancy of the final summary with a greedy
selection algorithm
\cite{cao2015ranking,yasunaga2017graph}. Redundancy is also addressed
in \citet{ganesan2010opinosis} who propose a graph-based framework for
abstractive summarization. \citet{wang2016neural} introduce an
encoder-decoder neural method for extractive opinion
summarization. Their approach requires direct supervision via
gold-standard extractive summaries for training, in contrast to our
weakly supervised formulation.

\section{Problem Formulation}
\label{sec:problem-formulation}

Let~$\mathrm{C}$ denote a corpus of reviews on a set of products
\mbox{$E_{\mathrm{C}} = \{e_i\}_{i=1}^{|E_{\mathrm{C}}|}$} from a
domain $d_{\mathrm{C}}$, e.g.,~televisions or keyboards. For every
product~$e$, the corpus contains a set of reviews \mbox{$R_e =
\{r_i\}_{i=1}^{|R_e|}$} expressing customers' opinions. Each review
$r_i$ is accompanied by the author's overall rating $y_i$ and is split
into segments $(s_1, \dots, s_m)$, where each segment $s_j$ is in turn
viewed as a sequence of words $(w_{j1}, \dots, w_{jn})$. A segment can
be a sentence, a phrase, or in our case an \textit{Elementary
Discourse Unit} (EDU; \citealt{mann1988rhetorical}) obtained from a
\textit{Rhetorical Structure Theory} (RST) parser \cite{feng2012text}.
EDUs roughly correspond to clauses and have been shown to facilitate
performance in summarization \cite{li2016role}, document-level
sentiment analysis \cite{bhatia2015better}, and single-document
opinion extraction \cite{angelidis18milnet}.

A segment may discuss zero or more \emph{aspects}, i.e.,~different
product attributes. We use \mbox{$A_{\mathrm{C}} = \{a_i\}_{i=1}^K$}
to refer to the aspects pertaining to domain $d_{\mathrm{C}}$. For
example, \textit{picture quality}, \textit{sound quality}, and
\textit{connectivity} are all aspects of televisions. By convention, a
\emph{general} aspect is assigned to segments that do not discuss any
specific aspects.  Let $A_s \subseteq A_{\mathrm{C}}$ denote the set
of aspects mentioned in segment~$s$; $\mathit{pol}_s \in [-1,+1]$
marks the \emph{polarity} a segment conveys, where~$-1$ indicates
maximally negative and $+1$~maximally positive sentiment. An opinion
is represented by tuple $o_s = (s, A_s, \mathit{pol}_s)$, and $O_e =
\{o_s\}_{s \in R_e}$ represents the set of all opinions expressed in
$R_e$.

For each product $e$, our goal is to produce a summary of the most
salient opinions expressed in reviews $R_e$, by selecting a small
subset \mbox{$S_e \subset O_e$}. We expect segments that discuss
specific product aspects to be better candidates for useful summaries.
We hypothesize that \emph{general} comments mostly describe customers'
overall experience, which can also be inferred by their 
rating, whereas aspect-related comments provide specific reasons for
their overall opinion.  We also assume that segments conveying highly positive
or negative sentiment are more likely to present informative opinions
compared to neutral ones, a claim supported by previous work
\cite{angelidis18milnet}.

We describe our novel approach to aspect extraction in Section
\ref{sec:aspects} and detail how we combine aspect, sentiment, and
redundancy information to produce opinion summaries in
Section~\ref{sec:opsumm}.

\section{Aspect Extraction}
\label{sec:aspects}

Our work builds on the aspect discovery model developed by
\citet{he2017unsupervised}, which we extend to facilitate the accurate
extraction of aspect-specific review segments in a more realistic setting.
In this section, we first describe their approach, point out its
shortcomings, and then present the extensions and modifications
introduced in our \textit{Multi-Seed Aspect Extractor} (MATE) model.

\subsection{Aspect-Based Autoencoder}
\label{sec:abae}

The \textit{Aspect-Based Autoencoder} (ABAE;
\citealt{he2017unsupervised}) is an adaptation of the
\textit{Relationship Modeling Network} \cite{iyyer2016feuding},
originally designed to identify attributes of fictional book
characters and their relationships. The model learns a segment-level
aspect predictor without supervision by attempting to reconstruct the
input segment's encoding as a linear combination of aspect embeddings.
ABAE starts by pairing each word $w$ with a pre-trained word embedding
\mbox{$\mathbf{v}_w \in \mathbb{R}^d$}, thus constructing a word
embedding dictionary~\mbox{$\mathbf{L} \in \mathbb{R}^{V \times d}$},
where V is the size of the vocabulary. The model also keeps an aspect
embedding dictionary \mbox{$\mathbf{A} \in \mathbb{R}^{K \times d}$},
where $K$ is the number of aspects to be identified and $i$-th row
$\mathbf{a}_i \in \mathbb{R}^d$ is a point in the word embedding
space.  Matrix $\mathbf{A}$ is initialized using the centroids from a
$k$-means clustering on the vocabulary's word embeddings.

The autoencoder, first produces a vector $\mathbf{v}_s$ for review
segment $s = (w_1, \dots, w_n)$ using an \textit{attention encoder}
that learns to attend on aspect words. A segment encoding is computed
as the weighted average of word vectors:
\begin{align}
  \mathbf{v}_s &= \sum_{i=1}^n c_i \mathbf{v}_{w_i} \label{eqn:segencoder} \\
  c_i &= \frac{exp(u_i)}{\sum_{j=1}^n exp(u_j)} \label{eqn:attention1} \\
  u_i &= \mathbf{v}_{w_i}^\mathsf{T} \cdot \mathbf{M} \cdot \mathbf{v}_s'\, ,
         \label{eqn:attention2}
\end{align}

\noindent where $c_i$ is the $i$-th word's attention weight,
$\mathbf{v}_s'$ is a simple average of the segment's word embeddings
and attention matrix $\mathbf{M} \in \mathbb{R}^{d \times d}$ is
learned during training.

Vector $\mathbf{v}_s$ is fed into a softmax classifier to predict a
probability distribution over~$K$ aspects:
\begin{equation}
  \mathbf{p}_s^{\mathit{asp}} = \softmax(\mathbf{W}\mathbf{v}_s + \mathbf{b})\, ,
  \label{eqn:asp}
\end{equation}
\noindent where $\mathbf{W} \in \mathbb{R}^{K \times d}$ and
$\mathbf{b} \in \mathbb{R}^K$ are the classifier's weight and bias
parameters.  The segment's vector is then reconstructed as the
weighted sum of aspect embeddings:
\begin{equation}
  \mathbf{r}_s = \mathbf{A}^\mathsf{T} \mathbf{p}_s^{\mathit{asp}}\, .
  \label{eqn:recon}
\end{equation}

The model is trained by minimizing a reconstruction loss
$J_r(\theta)$ that uses randomly sampled segments $n_1, n_2, \dots,
n_{k_n}$ as negative examples:\footnote{ABAE also uses a uniqueness
regularization term that is not shown here and is not used in our
Multi-Seed Aspect Extractor model.}
\begin{equation}
  J_r(\theta) = \sum_{s \in \mathrm{C}} \sum_{i=1}^{k_n} \mathrm{max}(0, 1 -
               \mathbf{r}_s \mathbf{v}_s + \mathbf{r}_s \mathbf{v}_{n_i}) 
  \label{eqn:loss}
\end{equation}

ABAE is essentially a neural topic model; it discovers topics which
will hopefully map to aspects, without any preconceptions about the
aspects themselves, a feature shared with most previous LDA-style
aspect extraction approaches
\cite{titov2008joint,he2017unsupervised,mukherjee-liu:2012:ACL20121}.
These models will set the number of topics to be discovered to a much
larger number ($\sim\!\!15$) than the actual aspects found in the data
($\sim\!5$). This requires a many-to-one mapping between discovered
topics and genuine aspects which is performed manually.

\begin{figure}[t]
  \centering
  \includegraphics[width=\columnwidth]{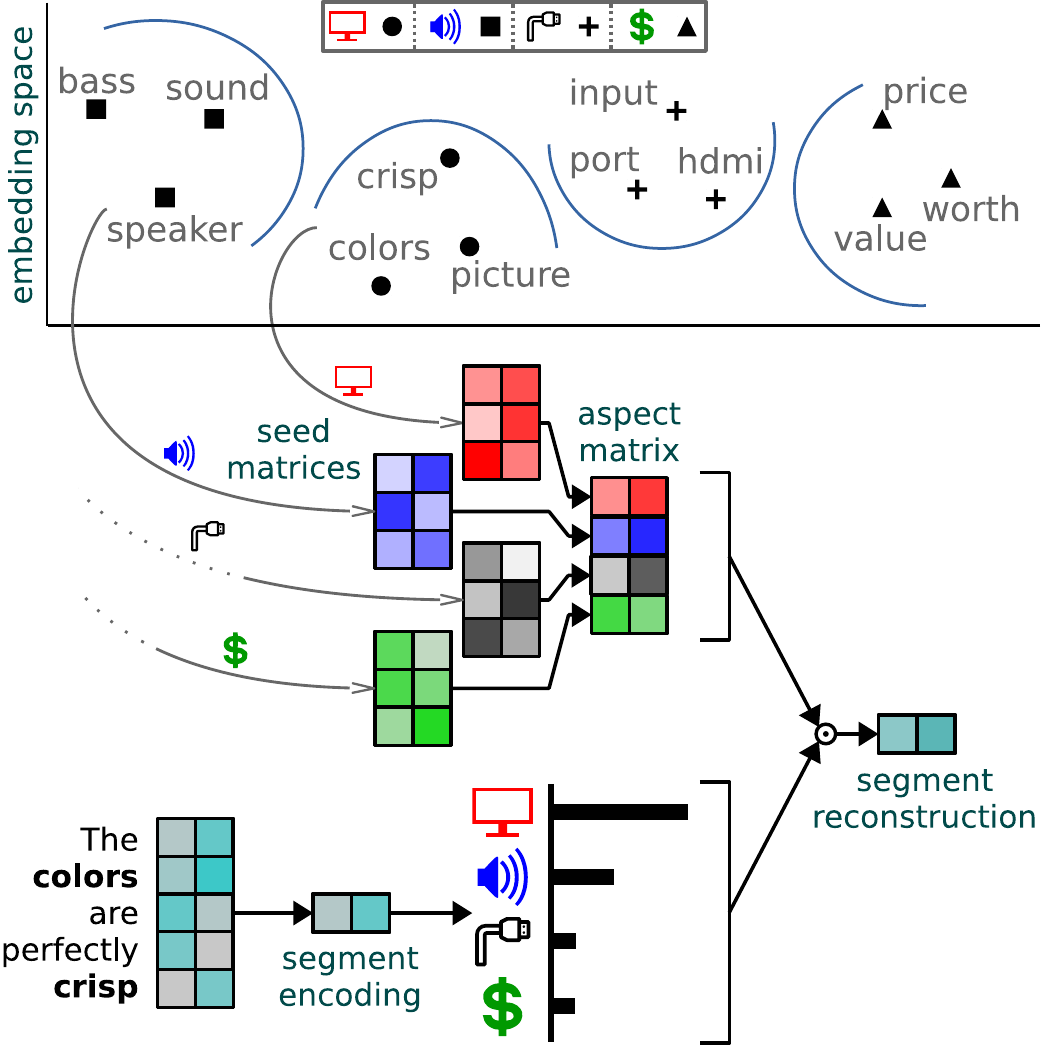}
  \caption{Multi-Seed Aspect Extractor (MATE).}
  \label{fig:autoencoder}
  \vspace{-1.5mm}
\end{figure}

\subsection{Multi-Seed Aspect Extractor}
\label{sec:msae}

Dynamic aspect extraction is advantageous since it assumes nothing
more than a set of relevant reviews for a product and may discover
unusual and interesting aspects (e.g.,~whether a plasma television has
protective packaging). However, it suffers from the fact that the
identified aspects are fine-grained, they have to be interpreted
post-hoc, and manually mapped to coarse-grained ones.

We propose a new weakly-supervised set-up for aspect extraction which
requires little human involvement. For every aspect $a_i \in
A_{\mathrm{C}}$, we assume there exists a small set of seed words
$\{\mathit{sw}_j\}_{j=1}^l$ which are good descriptors of~$a_i$.  We
can think of these \emph{seeds} as query terms that someone would use
to search for segments discussing $a_i$.  They can be set manually by
a domain expert or selected using a small number of aspect-annotated
reviews.  Figure~\ref{fig:autoencoder} (top) depicts four television
aspects (\textit{image}, \textit{sound}, \textit{connectivity} and
\textit{price}) and three of their seeds in word embedding space. MATE
replaces ABAE's aspect dictionary with multiple seed matrices
$\{\mathbf{A}_1, \mathbf{A}_2, \dots, \mathbf{A}_K\}$. Every matrix
$\mathbf{A}_i \in \mathbb{R}^{l \times d}$, contains one row per seed
word and holds the seeds' word embeddings, as illustrated by the set
of $[3 \times 2]$ matrices in Figure~\ref{fig:autoencoder}.

MATE still needs to produce an aspect matrix $\mathbf{A} \in
\mathbb{R}^{K \times d}$, in order to reconstruct the input segment's
embedding. We accomplish this by reducing each seed matrix to a single
aspect embedding with the help of seed weight vectors~$\mathbf{z}_i
\in \mathbb{R}^l$ ($\sum_j z_{ij} = 1$), and concatenating the
results, illustrated by the $[4 \times 2]$ aspect matrix in
Figure~\ref{fig:autoencoder}:
\begin{align}
  \mathbf{a}_i &= \mathbf{A}^\mathsf{T}_i \mathbf{z}_i \label{eqn:reduce} \\
  \mathbf{A} &= [\mathbf{a}^\mathsf{T}_1; \dots; \mathbf{a}^\mathsf{T}_K]\, . \label{eqn:concat}
\end{align}
The segment is reconstructed as in Equation~(\ref{eqn:recon}).  Weight
vectors $\mathbf{z}_i$ can be uniform (for manually selected seeds),
fixed, learned during training, or set dynamically for each input
segment, based on the cosine distance of its encoding to each seed
embedding. Our experiments showed that fixed weights, selected through
a technique described below, result in most stable performance
across domains. We only focus on this variant due to space
restrictions (but provide more details in the supplementary material).

When a small number of aspect-annotated reviews are available, seeds
and their fixed seed weights can be selected automatically. To obtain
a ranked list of terms that are most characteristic for each aspect,
we use a variant of the \textit{clarity} scoring function which was
first introduced in information retrieval
\cite{townsend2002predicting}. Clarity measures how much more likely
it is to observe word~$w$ in the subset of segments that discuss
aspect $a$, compared to the corpus as a whole:
\begin{equation}
  \mathrm{score}_a(w) = t_a(w) \, \mathrm{log}_2\frac{t_a(w)}{t(w)}\, ,
  \label{eqn:clarity}
\end{equation}
where $t_a(w)$ and $t(w)$ are the $l_1$-normalized
\mbox{\textit{tf-idf}} scores of~$w$ in the segments annotated with
aspect~$a$ and in all annotated segments, respectively.  Higher scores
indicate higher term importance and truncating the ranked list of
terms gives a fixed set of seed words, as well as their seed weights
by normalizing the scores to add up to one.  Table~\ref{tab:topwords}
shows the highest ranked terms obtained for every aspect in the
\textit{televisions} domain of our corpus (see Section
\ref{sec:oposum-dataset} for a detailed description of our data).

\subsection{Multi-Task Objective}
\label{sec:multitask}

MATE (and ABAE) relies on the attention encoder to identify and attend
to each segment's aspect-signalling words. The reconstruction
objective only provides a weak training signal, so we devise a
multi-task extension to enhance the encoder's effectiveness without
additional annotations.

\begin{table}[t]
  \small
\begin{center}
  \begin{tabular}{@{~}l@{~}@{~}l@{~}} \hline
  \textbf{Aspect} & \textbf{Top Terms} \\
  \hline
  Image                 & picture color quality black bright\\
  Sound                 & sound speaker quality bass loud\\
  Connectivity          & hdmi port computer input component \\
  Price                 & price value money worth paid\\
  Apps \& Interface     & netflix user file hulu apps\\
  Ease of Use           & easy remote setup user menu\\
  Customer Service~~~~~ & paid support service week replace\\
  Size \& Look          & size big bigger difference screen \\
  General               & tv bought hdtv happy problem\\ \hline
  \end{tabular}
\end{center}
  \caption{Highest ranked words for the television corpus according to
    Equation~(\ref{eqn:clarity}).}
  \label{tab:topwords}
  \vspace{-1.5mm}
\end{table}

We assume that aspect-relevant words not only provide a better basis
for the model's aspect-based reconstruction, but are also good
indicators of the product's domain.  For example, the words
\textsl{colors} and \textsl{crisp}, in the segment ``\textsl{The
colors are perfectly crisp}'' should be sufficient to infer that the
segment comes from a television review, whereas the words
\textsl{keys} and \textsl{type} in the segment ``\textsl{The keys feel
great to type on}'' are more representative of the keyboard domain.
Additionally, all four words are characteristic of specific
aspects.

Let $\mathrm{C}_{\mathit{all}} = \mathrm{C}_1 \cup \mathrm{C}_2 \cup
\dots$ denote the union of multiple review corpora, where
$\mathrm{C}_1$ is considered \textit{in-domain} and the rest are
considered \textit{out-of-domain}. We use $d_s \in \{d_{\mathrm{C}_1},
d_{\mathrm{C}_2}, \dots\}$ to denote the true domain of segment~$s$
and define a classifier that uses the vectors from our segment encoder
as inputs:
\begin{equation}
  \mathbf{p}_s^{\mathit{dom}} = \softmax(\mathbf{W}_\mathrm{C}\mathbf{v}_s + \mathbf{b}_\mathrm{C})\, ,
  \label{eqn:domain}
\end{equation}
\noindent where $\mathbf{p}_s^{\mathit{dom}} = \langle
p^{(d_{\mathrm{C}_1})}, p^{(d_{\mathrm{C}_2})}, \dots \rangle$ is a
probability distribution over product domains for segment~$s$
and $\mathbf{W}_\mathrm{C}$ and $\mathbf{b}_\mathrm{C}$ are the
classifier's weight and bias parameters. We use the negative log
likelihood of the domain prediction as the objective function,
combined with the reconstruction loss of Equation~(\ref{eqn:recon}) to
obtain a multi-task objective:

\begin{equation}
  J_{\mathrm{MT}}(\theta) = J_r(\theta) - \lambda \sum_{s \in \mathrm{C}_{\mathit{all}}}\mathrm{log} \, p^{(d_s)}\, ,
  \label{eqn:mtloss}
\end{equation}

\noindent where $\lambda$ controls the influence of the classification
loss. Note that the negative log-likelihood is summed over all
segments in~$\mathrm{C}_{\mathit{all}}$, whereas $J_r(\theta)$ is only
summed over the \mbox{in-domain} segments \mbox{$s \in \mathrm{C}_1$}.
It is important not to use the \mbox{out-of-domain} segments for
segment reconstruction, as they will confuse the aspect extractor due
to the aspect mismatch between different domains.

\begin{table*}
  \parbox{.41\textwidth}{
    \centering
    \small
    \begin{tabular}{@{~}lr@{~}} \hline
      \textbf{Segment} & \textbf{Salience} \\
      \hline
      1. The color and definition are perfect. &   \texttt{[+]}0.89\\
      2. Set up was extremely easy, &  \texttt{[+]}0.79\\
      3. Not worth \$ 300. &  \texttt{[-]}0.75\\
      4. The sound on this is horrendous. &  \texttt{[-]}0.52\\
      5. The sound is TERRIBLE. &  \texttt{[-]}0.45\\
      6. Nice and bright with good colors. &  \texttt{[+]}0.44\\\hline
    \end{tabular}
    \caption{Most salient opinions according to scores from
    Equation~\eqref{eqn:saliency} for an LCD TV.}
    \label{tab:ranking}
  \vspace{-1.5mm}
  }
  \hspace{1mm}
  \parbox{.54\textwidth}{
    \centering
    \small
    \begin{tabular}{@{~}lrrrr@{~}}\hline
      \textbf{Domain} & \multicolumn{1}{c}{\textbf{Products}} 
      & \multicolumn{1}{c}{\textbf{Reviews}}
      & \multicolumn{1}{c}{\textbf{EDUs}}
      & \textbf{Vocab}~ \\
      \hline
      Laptop Cases &  2,040 (10) &  42,727 (100) &  602K (1,262) & 30,443  \\
      B/T Headsets &  1,471 (10) &  80,239 (100) & 1.46M (1,344) & 51,263  \\
      Boots        &  4,723 (10) &  77,593 (100) &  987K (1,198) & 30,364  \\
      Keyboards    &    983 (10) &  33,713 (100) &  625K (1,396) & 34,095  \\
      Televisions  &  1,894 (10) &  56,510 (100) & 1.47M (1,483) & 59,051  \\
      Vacuums      &  1,184 (10) &  68,266 (100) & 1.50M (1,492) & 46,259  \\ \hline
      % TOTAL        & 12,295 (60) & 359,048 (600) & 6.65M (8,175) & 140,100 \\
    \end{tabular}
    \caption{The \textsc{OpoSum} corpus. Numbers in
      parentheses correspond to the human-annotated subset.}
    \label{tab:datastats}
  \vspace{-1.5mm}
  }
  \hfill
\end{table*}

\section{Opinion Summarization}
\label{sec:opsumm}

We now move on to describe our opinion summarization framework which
is based on the aspect extraction component discussed so far, a
polarity prediction model, and a segment selection policy which
identifies and discards redundant opinions.

\paragraph{Opinion Polarity}
\label{sec:polarity}

Aside from describing a product's aspects, segments also express
polarity (i.e.,~positive or negative sentiment). We identify segment
polarity with the recently proposed \textit{Multiple Instance Learning
  Network} model (\textsc{MilNet};
\citealt{angelidis18milnet}). Whilst trained on freely available
document-level sentiment labels, i.e., customer ratings on a scale
from 1 (negative) to 5 (positive), \textsc{MilNet} learns a
segment-level sentiment predictor using a hierarchical,
attention-based neural architecture.

Given review~$r$ consisting of segments $(s_1, \dots, s_m)$,
\textsc{MilNet} uses a CNN segment encoder to obtain segment vectors
$(\mathbf{u}_1, \dots, \mathbf{u}_m)$, each used as input to a
segment-level sentiment classifier.  For every vector $\mathbf{u}_i$,
the classifier produces a sentiment prediction
$\mathbf{p}_i^{\mathit{stm}} = \langle p_i^{(1)}, \dots, p_i^{(M)}
\rangle$, where $p_i^{(1)}$ and $p_i^{(M)}$ are probabilities assigned
to the most negative and most positive sentiment class respectively.
Resulting segment predictions $(\mathbf{p}_1^{\mathit{stm}}, \dots,
\mathbf{p}_m^{\mathit{stm}})$ are combined via a GRU-based attention
mechanism to produce a document-level prediction
$\mathbf{p}_r^{\mathit{stm}}$ and the model is trained end-to-end on
the reviews' user ratings using negative log-likelihood.

The essential by-product of \textsc{MilNet} are segment-level sentiment
predictions $\mathbf{p}_i^{\mathit{stm}}$, which are transformed into
polarities $\mathit{pol}_{s_i}$, by projecting them onto the~$[-1,
+1]$ range using a uniformly spaced sentiment class weight vector.

\paragraph{Opinion Ranking}
\label{sec:ranking}

Aspect predictions $\mathbf{p}_s^{\mathit{asp}} = \langle
p_s^{(a_1)},\dots, p_s^{(a_K)} \rangle$ and polarities
$\mathit{pol}_{s}$, form the opinion set $O_e = \{(s, A_s,
\mathit{pol}_s)\}_{s \in R_e}$ for every product $e \in
E_{\mathrm{C}}$. For simplicity, we set the predicted aspect-set $A_s$
to only include the aspect with the highest probability, although it
is straightforward to allow for multiple aspects. We rank every
opinion~$o_s \in O_e$ according to its salience:
\begin{equation}
  \mathit{sal}(o_s) = |\mathit{pol}_s| \cdot ( \max_i p_s^{(a_i)} -
  p_s^{(\mathit{\mathrm{GEN}})} )\, ,
  \label{eqn:saliency}
\end{equation}
\noindent where the quantity in parentheses is the probability
difference between the most probable aspect and the \textit{general}
aspect. The salience score will be high for opinions that are very
positive or very negative and are also likely to discuss a non-general
aspect.

\paragraph{Opinion Selection}
\label{sec:selection}

The final step towards producing summaries is to discard potentially
redundant opinions, something that is not taken into account by our
salience scoring method. Table \ref{tab:ranking} shows a partial
ranking of the most salient opinions found in the reviews for an LCD
television. All segments provide useful information, but it is evident
that segments 1 and 6 as well as 4 and 5 are paraphrases of the same
opinions.

We follow previous work on multi-document summarization
\cite{cao2015ranking,yasunaga2017graph} and use a greedy algorithm to
eliminate redundancy. We start with the highest ranked opinion, and
keep adding opinions to the final summary one by one, unless the
cosine similarity between the candidate segment and any segment
already included in the summary is lower than~$0.5$.

\begin{table*}
  \centering
  \small
  \begin{tabularx}{\textwidth}{lYYYYYYY}
    \hline \\[-3mm]
    \textbf{Aspect Extraction} (F1) & L. Bags & B/T H/S & Boots & Keyb/s & TVs & Vac/s & \textbf{AVG} \\[0.0mm]
    \hline \\[-3mm]
    Majority      & 37.9 & 39.8 & 37.1 & 43.2 & 41.7 & 41.6 & 40.2 \\
    ABAE          & 38.1 & 37.6 & 35.2 & 38.6 & 39.5 & 38.1 & 37.9 \\
    ABAE$_{init}$ & 41.6 & 48.5 & 41.2 & 41.3 & 45.7 & 40.6 & 43.2 \\
    MATE          & 46.2 & 52.2 & 45.6 & 43.5 & 48.8 & 42.3 & 46.4 \\
    MATE+MT       & \textbf{48.6} & \textbf{54.5} & \textbf{46.4}
                  & \textbf{45.3} & \textbf{51.8} & \textbf{47.7} 
                  & \textbf{49.1} \\

    \hline \\[-1mm] \hline \\[-3mm]
    \textbf{ Salience} (MAP/P@5) & L. Bags & B/T H/S & Boots & Keyb/s & TVs & Vac/s & \textbf{AVG} \\[0.0mm] 
    \hline \\[-3mm]
    \milnet                  & 21.8 / 40.0 & 19.8 / 36.7 & 17.0 / 39.3 & 14.1 / 28.0 & 14.3 / 36.0 & 14.6 / 31.3 & 16.9 / 35.2 \\
    ABAE$_{init}$ & 19.9 / 48.5 & 27.5 / 49.7 & 13.8 / 28.1 & 19.0 / 44.9 & 16.8 / 42.4 & 16.1 / 34.0 & 18.8 / 41.3 \\
    MATE          & 23.0 / 57.1 & 30.9 / 50.7 & 15.4 / 31.9 & 21.0 / 43.1 & 18.7 / 44.7 & 19.9 / 44.0 & 21.5 / 45.2 \\
    MATE+MT       & 26.3 / 60.8 & 37.5 / 66.7 & 17.3 / 33.6 & 20.9 / 44.9 & 23.6 / 48.0 & 22.4 / 43.9 & 24.7 / 49.6 \\
    \milnet+ABAE$_{init}$    & 27.1 / 56.0 & 33.5 / 66.5 & 19.3 / 34.8 & 22.4 / 51.7 & 19.0 / 43.7 & 20.8 / 43.5 & 23.7 / 49.4 \\
    \milnet+MATE             & 28.2 / 54.7 & 36.0 / 66.5 & 21.7 / 39.3 & 24.0 / 52.0 & 20.8 / 46.1 & 23.5 / 49.3 & 25.7 / 51.3 \\
    \milnet+MATE+MT          & \textbf{32.1} / \textbf{69.2} & \textbf{40.0} / \textbf{74.7}
                             & \textbf{23.3} / \textbf{40.4} & \textbf{24.8} / \textbf{56.4}
                             & \textbf{23.8} / \textbf{52.8} & \textbf{26.0} / \textbf{53.1}
                             & \textbf{28.3} / \textbf{57.8} \\
    \hline
  \end{tabularx}
  \caption{Experimental results for the 
    identification of aspect segments (top) and the retrieval of
    salient segments (bottom) on \textsc{OpoSum}'s six product domains and
    overall (AVG).}
  \vspace{-1.5mm}
  \label{tab:results}
\end{table*}

\section{The \textsc{OpoSum} Dataset}
\label{sec:oposum-dataset}

We created \textsc{OpoSum}, a new dataset for the training and
evaluation of \textbf{Op}ini\textbf{o}n \textbf{Sum}marization models
which contains Amazon reviews from six product domains: \textit{Laptop
  Bags}, \textit{Bluetooth Headsets}, \textit{Boots},
\textit{Keyboards}, \textit{Televisions}, and \textit{Vacuums}. The
six training collections were created by downsampling from the
\textit{Amazon Product
  Dataset}\footnote{\url{http://jmcauley.ucsd.edu/data/amazon/}}
introduced in \citet{mcauley2015image} and contain reviews and their
respective ratings. The reviews were segmented into EDUs using a
publicly available RST parser \cite{feng2012text}.

To evaluate our methods and facilitate research, we produced a
human-annotated subset of the dataset. For each domain, we uniformly
sampled (across ratings) 10 different products with 10 reviews each,
amounting to a total of 600 reviews, to be used only for development
(300) and testing (300). We obtained EDU-level aspect annotations,
salience labels and gold standard opinion summaries, as described
below. Statistics are provided in Table~\ref{tab:datastats} and in 
supplementary material.

\paragraph{Aspects} For every domain, we pre-selected nine
representative aspects, including the \textit{general} aspect. We
presented the EDU-segmented reviews to three annotators and asked them
to select the aspects discussed in each segment (multiple aspects were
allowed).  Final labels were obtained using a majority vote among
annotators.  Inter-annotator agreement across domains and annotated
segments using Cohen's Kappa coefficient was $K = 0.61$ ($N =
8,\!175$, $k=3$).

\paragraph{Opinion Summaries} We produced opinion summaries for the 60
products in our benchmark using a two-stage procedure. First, all
reviews for a product were shown to three annotators. Each annotator
read the reviews one-by-one and selected the subset of segments they
thought best captured the most important and useful comments, without
taking redundancy into account.  This phase produced binary
\textit{salience} labels against which we can judge the ability of a
system to identify important opinions. Again, using the Kappa
coefficient, agreement among annotators was \mbox{$K = 0.51$}
($N=8,\!175$, $k=3$).\footnote{While this may seem moderate,
\citet{radev2003evaluation} show that inter-annotator agreement for
extractive summarization is usually lower ($K<0.30$).} In the second
stage, annotators were shown the salient segments they identified (for
every product) and asked to create a final extractive summary by
choosing opinions based on their popularity, fluency and clarity,
while avoiding redundancy and staying under a budget of 100 words. We
used ROUGE \cite{lin2003automatic} as a proxy to inter-annotator
agreement. For every product, we treated one reference summary as
system output and computed how it agrees with the rest.  ROUGE scores
are reported in Table~\ref{tab:rouge} (last row).

\section{Experiments}
\label{sec:experiments}

In this section, we discuss implementation details and present our
experimental setup and results. We evaluate model performance on three
subtasks: aspect identification, salient opinion extraction, and
summary generation.

\paragraph{Implementation Details}
Reviews were lemmatized and stop words were removed. We initialized
MATE using \mbox{200-dimensional} word embeddings trained on each
product domain using skip-gram \cite{mikolov2013distributed} with
default parameters.  We used 30~seed words per aspect, obtained via
Equation~(\ref{eqn:clarity}).  Word embeddings $\mathbf{L}$, seed
matrices $\{\mathbf{A}_i\}_{i=1}^K$ and seed weight vectors
$\{\mathbf{z}_i\}_{i=1}^K$ were fixed throughout training. We used the
Adam optimizer \cite{kingma2014adam} with learning rate $10^{-4}$ and
mini-batch size~50, and trained for 10~epochs.  We used 20~negative
examples per input for the reconstruction loss and, when used, the
multi-tasking coefficient $\lambda$~was set to~10. Seed words and
hyperparameters were selected on the development set and we report
results on the test set, averaged over~5 runs.

\paragraph{Aspect Extraction}
We trained aspect models on the collections of
Table~\ref{tab:datastats} and evaluated their predictions against the
human-annotated portion of each corpus. Our MATE model and its
multi-task counterpart (MATE+MT) were compared against a majority
baseline and two ABAE variants: vanilla ABAE, where aspect matrix
$\mathbf{A}$ is initialized using $k$-means centroids and fine-tuned
during training; and ABAE$_{init}$, where rows of $\mathbf{A}$ are
fixed to the centroids of respective seed embeddings. This allows us
to examine the benefits of our multi-seed aspect representation.
Table \ref{tab:results} (top) reports the results using micro-averaged
F1.  Our models outperform both variants of ABAE across domains.
ABAE$_{init}$ improves upon the vanilla model, affirming that informed
aspect initialization can facilitate the task. The richer multi-seed
representation of MATE, however, helps our model achieve a
3.2\%~increase in F1. Further improvements are gained by the
multi-task model, which boosts performance by~2.7\%.

\paragraph{Opinion Salience} We are also interested in our system's
ability to identify salient opinions in reviews. The first phase of
our opinion extraction annotation provides us with binary salience
labels, which we use as gold standard to evaluate system opinion
rankings. For every product~$e$, we score each segment $s \in R_e$
using Equation~(\ref{eqn:saliency}) and evaluate the obtained rankings
via Mean Average Precision (MAP) and Precision at the 5th retrieved
segment (P@5).\footnote{A system's salience ranking is individually
compared against labels from each annotator and we report the
average.}  Polarity scores were produced via \milnet{}; we obtained
aspect probabilities from ABAE$_{init}$, MATE, and MATE+MT. We also
experimented with a variant that only uses \milnet's polarities and,
additionally, with variants that ignore polarities and only use aspect
probabilities.

Results are shown in Table \ref{tab:results} (bottom). The combined
use of polarity and aspect information improves the retrieval of
salient opinions across domains, as all model variants that use our
salience formula of Equation~(\ref{eqn:saliency}) outperform the
\milnet{}- and aspect-only baselines. When comparing between
aspect-based alternatives, we observe that the extraction accuracy
correlates with the quality of aspect prediction. In particular,
ranking using \milnet{}+MATE+MT gives best results, with a
2.6\%~increase in MAP against \milnet{}+MATE and 4.6\%~against
\milnet{}+ABAE$_{init}$. The trend persists even when \milnet{}
polarities are ignored, although the quality of rankings is worse in
this case.

\begin{table}
  \centering
  \small
  \begin{tabularx}{\columnwidth}{lYYY}
    \hline \\[-3mm]
    \textbf{Summarization} & \multicolumn{3}{c}{\scriptsize ROUGE-1~~~ROUGE-2~~~ROUGE-L} \\
    \hline \\[-3mm]
    Random             & ~35.1 & 11.3 & 34.3 \\
    Lead               & ~35.5 & 15.2 & 34.8 \\
    SumBasic           & ~34.0 & 11.2 & 32.6 \\
    LexRank            & ~37.7 & 14.1 & 36.6 \\
    Opinosis           & ~36.8 & 14.3 & 35.7 \\
    Opinosis+MATE+MT   & ~38.7 & 15.8 & 37.4 \\
    \milnet+MATE+MT    & ~43.5 & 21.7 & 42.8 \\
    \milnet+MATE+MT+RD & ~\textbf{44.1} & \textbf{21.8} &
    \textbf{43.3} \\
Inter-annotator Agreement & ~54.7 & 36.6 & 53.9\\
    \hline
  \end{tabularx}
  \vspace{-1.5mm}
  \caption{Summarization results on \textsc{OpoSum}.}
  \label{tab:rouge}
\end{table}

\paragraph{Opinion Summaries}
We now turn to the summarization task itself, where we compare our
best performing model (\milnet+MATE+MT), with and without a redundancy
filter (RD), against the following methods: a baseline that
selects~segments \textit{randomly}; a \textit{Lead} baseline that only
selects the leading segments from each review; \textit{SumBasic}, a
generic frequency-based extractive summarizer
\cite{nenkova2005impact}; \textit{LexRank}, a generic graph-based
extractive summarizer \cite{erkan2004lexrank}; \textit{Opinosis}, a
graph-based abstractive summarizer that is designed for opinion
summarization \cite{ganesan2010opinosis}. All extractive methods
operate on the EDU level with a \mbox{100-word} budget. For Opinosis,
we tested an aspect-agnostic variant that takes every review segment
for a product as input, and a variant that uses MATE's groupings of
segments to produce and concatenate aspect-specific summaries.

\begin{table}
  \centering
  \small
  \begin{tabular}{lcccc}
    \hline \\[-3mm]
    & Inform. & Polarity & Coherence & Redund. \\
    \hline \\[-3mm]
    Gold      & ~2.04          & \textbf{~8.70} & \textbf{10.93}~ & \textbf{~6.11} \\
    This work & \textbf{~9.26} & ~3.15          & ~1.11           & ~2.96 \\
    Opinosis  &-12.78~~        & -10.00~~       & -9.08           & -9.45 \\
    Lead      & ~1.48          & -1.85          & -2.96           & ~0.37 \\
    \hline
  \end{tabular}
  \vspace{-1.5mm}
  \caption{\textit{Best-Worst Scaling} human evaluation.}
  \vspace{-1.5mm}
  \label{tab:human}
\end{table}

Table~\ref{tab:rouge} presents ROUGE-1, ROUGE-2 and ROUGE-L F1 scores,
averaged across domains. Our model (\milnet+MATE+MT) significantly
outperforms all comparison systems ($p<0.05$; paired bootstrap
resampling; \citealt{koehn:2004:EMNLP}), whilst using a redundancy
filter slightly improves performance. Assisting Opinosis with aspect
predictions is beneficial, however, it remains significantly inferior
to our model (see the supplementary material for additional results).

\begin{figure*}[t]
\begin{small}
\begin{tabular}{cp{14.9cm}}\hline \\[-3mm]
\multicolumn{2}{l}{\textbf{Product domain:} Televisions} \\
\multicolumn{2}{l}{\textbf{Product name:} Sony BRAVIA 46-Inch HDTV}\\ \hline
\parbox[t]{1mm}{\multirow{4}{*}{\rotatebox[origin=c]{90}{\small \textbf{Human}}}}
  &  Plenty of ports and settings.
  Easy hookups to audio and satellite sources.
  The sound is good and strong.
  This TV looks very good.
  and the price is even better.
  The on-screen menu/options is quite nice.
  and the internet apps work as expected.
  The picture is clear and sharp.
  which is TOO SLOW to stream HD video... 
  The software and apps built into this TV.
  are difficult to use and setup.
  Their service is handled off shore making.
  communication a bit difficult. :( \\ \hline
\parbox[t]{1mm}{\multirow{5}{*}{\rotatebox[origin=c]{90}{\small \textbf{LexRank}}}}
  &  Get a Roku or Netflix box.
  I watch cable, Netflix, Hulu Plus, YouTube videos and computer movie
  files on it.
  Sound is good much better.
  DO NOT BUY! this SONY Bravia ` Smart ' TV...
  and avoid the Sony apps at all costs.
  Because of these two issues, I returned the Sony TV.
  Also you can change the display and sound settings on each port.
  However, the streaming speed for netflix is just down right terrible.
  Most of the time I just quit.
  Since I do not own the cable box,
  So, I have the cable.\\ \hline
\parbox[t]{1mm}{\multirow{5}{*}{\rotatebox[origin=c]{90}{\small \textbf{Opinosis}}}}
  & The picture and not bright at all even compared to my
  6-year old sony lcd tv. It will not work with an hdmi. 
  Connection because of a conflict with comcast's dhcp. 
  Being generous because I usuallly like the design and
  attention to detail of sony products). I am very
  disappointed with this tv for two reasons: picture
  brightness and channel menu. Numbers of options available 
  in the on-line area of the tv are numerous and extremely
  useful. Wow look at the color, look at the sharpness of 
  the picture, amazing and the amazing. \\\hline
\parbox[t]{1mm}{\multirow{3}{*}{\rotatebox[origin=c]{90}{\small \textbf{This work}}}}
  & 
  Plenty of ports and settings
  and have been extremely happy with it.
  The sound is good and strong.
  The picture is beautiful.
  And the internet apps work as expected.
  And the price is even better.
  Unbelieveable picture and the setup is so easy.
  Wow look at the color, look at the sharpness of the picture.
  The Yahoo! widgets do not work.
  And avoid the Sony apps at all costs. 
  Communication a bit difficult. :( \\ \hline
\end{tabular}
\end{small} 
\vspace{-0.5mm}
\caption{Human and system summaries for a product in the
    \textit{Televisions} domain.}
  \label{tab:tv-example}
\end{figure*}

We also performed a large-scale user study. For every product
in the \textsc{OpoSum} test set, participants were asked to compare
summaries produced by: a (randomly selected) human annotator, our best
performing model (\milnet+MATE+MT+RD), Opinosis, and the Lead
baseline. The study was conducted on the Crowdflower platform using
\textit{Best-Worst Scaling} (BWS; \citealt{louviere1991best,
louviere2015best}), a less labour-intensive alternative to paired
comparisons that has been shown to produce more reliable results than
rating scales \cite{kiritchenko2017best}. We arranged every 4-tuple of
competing summaries into four triplets. Every triplet was shown to
three crowdworkers, who were asked to decide which summary was
\textit{best} and which one was \textit{worst} according to four
criteria: \textit{Informativeness} (How much useful information about
the product does the summary provide?), \textit{Polarity} (How well
does the summary highlight positive and negative opinions?),
\textit{Coherence} (How coherent and easy to read is the summary?)
\textit{Redundancy} (How successfully does the summary avoid redundant
opinions?).

For every criterion, a system's score is computed as the percentage of
times it was selected as best minus the percentage of times it was
selected as worst \cite{orme2009maxdiff}. The scores range from -100
(unanimously worst) to +100 (unanimously best) and are shown in
Table~\ref{tab:human}. Participants favored our model over comparison
systems across all criteria (all differences are statistically
significant at $p<0.05$ using post-hoc HD Tukey tests). Human
summaries are generally preferred over our model, however the
difference is significant only in terms of coherence ($p<0.05$).

Finally, Figure~\ref{tab:tv-example} shows example summaries for a
product from our televisions domain, produced by one of our annotators
and by 3 comparison systems (LexRank, Opinosis and our
\milnet+MATE+MT+RD). The human summary is primarily focused on
aspect-relevant opinions, a characteristic that is also captured to a
large extent by our method. There is substantial overlap between
extracted segments, although our redundancy filter fails to identify a
few highly similar opinions (e.g., those relating to the picture
quality). The LexRank summary is inferior as it only identifies a few
useful opinions, and instead selects many general or non-opinionated
comments. Lastly, the abstractive summary of Opinosis does a good job
of capturing opinions about specific aspects but lacks in fluency, as
it produces grammatical errors. For additional system outputs, see
supplementary material.

\section{Conclusions}
We presented a weakly supervised neural framework for aspect-based
opinion summarization. Our method combined a seeded aspect extractor
that is trained under a multi-task objective without direct
supervision, and a multiple instance learning sentiment predictor, to
identify and extract useful comments in product reviews. We evaluated
our weakly supervised models on a new opinion summarization corpus
across three subtasks, namely aspect identification, salient opinion
extraction, and summary generation. Our approach delivered significant
improvements over strong baselines in each of the subtasks, while a
large-scale judgment elicitation study showed that crowdworkers favor
our summarizer over competitive extractive and abstractive systems.

In the future, we plan to develop a more integrated approach where
aspects and sentiment orientation are jointly identified, and work
with additional languages and domains. We would also like to
develop methods for abstractive opinion summarization using weak
supervision signals.

\paragraph{Acknowledgments} We gratefully acknowledge the financial
support of the European Research Council (award number 681760).

\nocite{zhao2005hierarchical}
%\nocite{radev2003evaluation}
\bibliography{emnlp2018}
\bibliographystyle{acl_natbib_nourl}

\end{document}